\documentclass[11pt,journal,letterpaper,twocolumn]{IEEEtran}

\usepackage{indentfirst}
\usepackage{amsfonts}
\usepackage{amssymb}
\usepackage{algorithmic}
\usepackage{algorithm}
\usepackage{cite}
\usepackage{graphicx}
\usepackage{psfrag}
\usepackage{amsmath,multirow}
\usepackage{color}

\newtheorem{definition}{Definition}[section]
\newtheorem{remark}{Remark}[section]
\newcommand{\R}{{\mathbb R}}
\newcommand{\be}{\begin{eqnarray}}
\newcommand{\ben}{\begin{eqnarray*}}
\newcommand{\en}{\end{eqnarray}}
\newcommand{\enn}{\end{eqnarray*}}

\begin{document}
\date{}




\title{Intrinsic dimension estimation of data by principal component analysis}
\author{Mingyu Fan\thanks{M. Fan and B. Zhang are with LSEC and the Institute
of Applied Mathematics, AMSS, Chinese Academy of Sciences, Beijing 100190,
China ~(email: fanmingyu@amss.ac.cn, b.zhang@amt.ac.cn)},
Nannan Gu\thanks{N. Gu and H. Qiao are with the Institute of Automation,
Chinese Academy of Sciences, Beijing 100190, China
(email: gunannan@gmail.com, hong.qiao@ia.ac.cn)}, Hong Qiao
and Bo Zhang\thanks{\bf Corresponding author: Bo Zhang}
}





\maketitle

\begin{abstract}
Estimating intrinsic dimensionality of data is a classic problem in
pattern recognition and statistics. Principal Component Analysis
(PCA) is a powerful tool in discovering dimensionality of data sets with a linear
structure; it, however, becomes ineffective when data have a nonlinear structure.
In this paper, we propose a new PCA-based method to estimate intrinsic
dimension of data with nonlinear structures.
Our method works by first finding a minimal cover of the data set,
then performing PCA locally on each subset in the cover and finally
giving the estimation result by checking up the data variance on all small
neighborhood regions.
The proposed method utilizes the whole data set to estimate its intrinsic
dimension and is convenient for incremental learning.
In addition, our new PCA procedure can filter out noise in data and converge to a
stable estimation with the neighborhood region size increasing.
Experiments on synthetic and real world data sets show effectiveness
of the proposed method.
\end{abstract}


\section{Introduction}

Intrinsic dimensionality (ID) of data is a key priori knowledge in
pattern recognition and statistics, such as time series analysis,
classification and neural networks, to improve their performance.
In time series analysis \cite{Timeseries1995}, the domain of attraction
of a nonlinear dynamic system has a very complex geometric
structure, and study on the geometry of the attraction domain is
closely related to the fractal geometry. Fractal dimension is an important tool
to characterize certain geometric properties of complex sets.
In neural network design \cite{Bishop1995}, the number of hidden units in the
encoding middle layer should be chosen according to the ID of data.
In classification tasks \cite{statistics}, in order to balance the
generalization ability and the empirical risk value, the complexity of the
function should also be related to the ID of data.

Recently, manifold learning, an important approach for nonlinear
dimensionality reduction, has drawn great interests. Important
manifold learning algorithms include isometric feature mapping
(Isomap) \cite{T2000}, locally linear embedding (LLE) \cite{R2000} and
Laplacian eigenmaps (LE) \cite{Lap2003}. They all assume data to
distribute on an intrinsically low-dimensional
sub-manifold \cite{SciencemaniPerception} and reduce the
dimensionality of data by investigating the intrinsic structure of data.
However, all manifold learning algorithms require the ID of
data as a key parameter for implementation.

Previous ID estimation methods can be categorized mainly into three
groups: projection approach, geometric approach and probabilistic approach.
The projection approach \cite{LocalPCA1971,KernelPCA1988,OTPMs1998} finds ID
by checking up the low-dimensional embedding of data.
The geometric method \cite{CamastraSurvey2003} finds ID by investigating the
intrinsic geometric structure of data.
The probabilistic technique \cite{MLE2004} builds estimators by making distribution
assumptions on data.
These approaches will be briefly introduced in Section \ref{previous}.

In this paper, we propose a new PCA-based method for ID estimation which is called the C-PCA method.
The proposed method first finds a minimal cover of the data set, and each subset in the cover
is considered as a small subregion of the data manifold. Then, on each subset, a revised PCA procedure
is applied to examine the local structure.
The revised PCA method can filter out noise in data and leads to a stable and convergent ID estimation
with the increase of the subregion size, as shown by the experimental results.
This is an advantage over the traditional PCA method which is very sensitive to noise, outliers and
the choice of the subregion size.
Further analysis shows that the revised PCA procedure can efficiently reduce
the running time complexity and utilize all data samples for ID estimation.
We should remark that our ID estimation method is also applicable to incremental learning
for consecutive data. Our method is compared with the maximum likelihood estimation (MLE)
method \cite{MLE2004}, the manifold adaptive method (which is referred to as
the $k$-$k/2$ NN method in this paper) \cite{manifoldAdapt}
and the k-nearest neighbor graph ($k$-NNG) method \cite{GeodesicEntropic2004,KNNgraphs2005}
through experiments.

The rest of the paper is organized as follows. In Section \ref{previous},
previous ID estimation methods are briefly reviewed. In Section \ref{C-PCA},
the new ID estimation method (C-PCA) is introduced. In Section \ref{exp},
experiments are conducted on synthetic and real world data sets to
show the effectiveness of the proposed algorithm.
Conclusion is made in Section \ref{con}.

\section{Previous algorithms on ID estimation}\label{previous}

Previously, there are mainly three approaches to estimate the ID of
data: projection, geometric and probabilistic approaches.

The projection approach first projects data into a low-dimensional space and then
determine the ID by verifying the low-dimensional representation of data.
PCA is a classical projection method which finds ID by counting the number of
significant eigenvalues. However, the traditional PCA only works on data lying in
a linear subspace but becomes ineffective when data distribute on a nonlinear manifold.
To overcome this limitation, local-PCA \cite{LocalPCA1971} and OTPMs PCA \cite{OTPMs1998}
have been proposed and can discover the ID of data lying on nonlinear
manifolds by performing the PCA method locally.
The Isomap algorithm yields ID of data by inspecting the elbow of residual variance
curve \cite{T2000}. Cheng et al. gave an efficient procedure to compute eigenvalues and
eigenvectors in PCA \cite{ProvablePCA2005}.

Geometric approaches make use of the geometric structure of data to build ID estimators.
Fractal-based methods have been well developed and used in time series analysis.
For example, the correlation dimension (a kind of fractal dimensions) was used in \cite{GP1983}
to estimate the ID, whilst the method of packing numbers was proposed in \cite{PackingDim2002}
to find the ID.
Other fractal-based methods include the kernel correlation method \cite{kernelCorrelation2005}
and the quantization estimator \cite{VectorQuantization}. A good survey on fractal-based methods
can be found in \cite{CamastraSurvey2003}.
There are also many methods based on techniques from computational geometry.
Lin \cite{RML2008} and Cheng \cite{Slivers2009} suggested to construct simplices to find the ID,
while the nearest neighbor approach uses the distances between data points with their
nearest neighbors to build ID estimators such as the estimator proposed
by Pettis et al. \cite{Petti1971}, the $k$-NNG method \cite{GeodesicEntropic2004,KNNgraphs2005}
and the incising ball method \cite{incisingball}.
A comparison of the local-PCA method with that introduced by Pettis et al. was made
in \cite{Evaluation1995}.

Probabilistic methods are based on probabilistic assumptions of data and have been tested
on various data sets with stable performance.
The MLE-method \cite{MLE2004} is a representative method of this approach, whose final
global estimator is given by averaging the local estimators:
\ben
\hat{d}_k(x_i)=\left[\frac{1}{k-1}\sum_{j=1}^{k-1}
\log\frac{T_k(x_i)}{T_j(x_i)}\right]^{-1}, \text{ for } i=1,\cdots,N,
\enn
where $T_k(x_i)$ is the distance between $x_i$ and its $k$-th
nearest neighbor. MacKay and Ghahramani \cite{revisedMLE} pointed
out that compared with averaging the local estimators directly, it is
more sensible to average their inverses $\hat{d}_k^{-1}(x_i)$,
$i=1,\cdots, N$ for the maximum likelihood purpose. The recommended
final estimator is
\ben
\hat{d}_k^{-1} =\frac{1}{N} \sum_{i=1}^N \hat{d}_k^{-1}(x_i),
\enn
where $d_k$ is the estimated ID corresponding to the neighborhood size $k$.

\section{ID estimation using PCA with cover sets: C-PCA}\label{C-PCA}

Basically, there are two kinds of definitions of ID that are
commonly used. One is based on the fractal dimension, such as the Hausdorff
dimension and the packing dimension that are usually real positive numbers.
The other kind of definition is based on the embedding manifold
whose ID is always an integer.

\begin{definition}[Embedding manifold and dimension]
Let $d<D$ and let $\Omega$ be a compact open set in $\R^d$. Assume that
$\mbox{span}\,\{\Omega-\int_{\Omega} d\mu\}=\R^d$ and $\phi:\Omega\to\R^D$
is a smooth function. The set $\mathcal{X}=\phi(\Omega)$ is
called an embedding manifold with $d$ its embedding dimension.
\end{definition}

More and more real world data are proved to have nonlinear intrinsic structures and
may possibly distribute on nonlinear embedding manifolds \cite{SciencemaniPerception}.
Therefore, estimation of embedding ID of data becomes an important problem \cite{incisingball}.
In this paper, we focus on estimation of embedding dimensions.

\subsection{PCA-based methods for ID estimation}\label{PCAdim}

The traditional PCA can find a subspace on which data projections have maximum variance.
Given a data set $\mathcal{X}=\{x_1,\cdots,x_N\}$ with $x_i \in \R^D$.
Let $X=[x_1,\cdots,x_N]$ and $\bar{x}=\frac{1}{N}\sum_{i=1}^N x_i$.
The covariance matrix of $\mathcal{X}$ is given by
\ben
C = \frac{1}{N}\sum_{i=1}^N (x_i-\bar{x})(x_i-\bar{x})^T.  \label{e1}
\enn
Since $C$ is a positive semi-definite matrix, we can assume that
$\lambda_1\geq\lambda_2\geq\cdots\geq\lambda_N\geq 0$ are the eigenvalues of $C$
with $\nu_1,\cdots, \nu_N$ the corresponding orthonormal eigenvectors, respectively.
The eigen-decomposition of matrix $C$ is denoted as
$C=\Gamma D\Gamma^T,$
where $D$ is a diagonal matrix with $D_{ii}=\lambda_i$ and $\Gamma=[\nu_1,\cdots,\nu_N]$.
The eigenvector $\nu_i$ is the $i$-th principal direction (PD) and, for any variable $x$,
$y_i= \nu_i x$ is defined as the $i$-th principal component (PC). By the definition,
we have the variance $\text{var}(y_i) =\lambda_i$ and the covariance $\text{cov}(y_i,y_j)=0$.

If the data set $\mathcal{X}$ distributes on a linear subspace, then the $d$ primary PDs
should be able to span the subspace and the corresponding PCs can account for most of the
variations contained in $\mathcal{X}$. On the other hand, the variance of PCs on
$\nu_{d+1},\cdots, \nu_N$ (i.e., the PDs which are orthogonal to the linear subspace of
dimension $d$) will be trivial. The most commonly-used criterions for ID estimation with
the PCA method are
\be
\frac{\underset{i=1,\cdots,d}\min\left(\text{var}(y_i)\right)}
{\underset{j=d+1,\cdots,N}\max\left(\text{var}(y_j)\right)}>\alpha\gg1 \label{e2}
\en
and the percentage of the accounted variance
\be\label{e3}
\frac{\sum_{i=1}^d\text{var}(y_i)}{\sum_{i=1}^N\text{var}(y_i)}>\beta,
\qquad 0<\beta<1.
\en
In this paper, the ID, $d$, is determined if the condition (\ref{e2}) or (\ref{e3}) is satisfied.

\subsection{Filtering out the noise of data}\label{PCAdim2}

There are two challenges for PCA-based ID estimation methods. The first one is how to
filter out the noise in data, while the second one is how to choose the size of subregions
on the manifold. Previously, the ID estimation of data obtained with PCA-based methods
always increases with the size of subregions so the methods can not converge to give a
stable ID estimation. In order to address these two limitations, we propose the following
noise filtering procedure which can efficiently filter out the noise in data
and make PCA-based methods to converge.

Consider the effect of additive white noise $\mu$ in data with $E(\mu)=0$ and
$\text{var}(\mu)=\sigma^2$. The covariance matrix of the noise corrupted data
is given by
\ben
C'=\text{var}(X+\mu)=C+\sigma^2 I,
\enn
where $C$ is the covariance matrix of the data $\mathcal{X}$.
It can be seen that the PDs of $C'$ are identical to those of $C$ and
the eigenvalues of $C'$ are $\lambda_i'=\lambda_i+\sigma^2.$
If $\sigma$ is relatively large, then the ID criterions (\ref{e2}) and
(\ref{e3}) will be ineffective.

The variance of data projections on the PDs that are orthogonal to the intrinsic
embedding subspace is very small, and the most part of the variance is produced
by noise. Therefore, it is possible to calculate the variance of noise
by projecting data on the orthogonal PDs.
Given a real number $P$ which is very close to $1$ ($P$ is taken to be $0.95$
in this paper), the noise part of data is determined by
\ben
\frac{\sum_{i=1}^{r-1}\text{var}(y_i)}{\sum_{i=1}^N\text{var}(y_i)}<P\quad
\text{and }\quad\frac{\sum_{i=1}^r\text{var}(y_i)}{\sum_{i=1}^N\text{var}(y_i)}>P.
\label{e4}
\enn
Thus, the variance of noise contained in data can be estimated as
\be
\hat{\sigma}^2 = \frac{1}{N-r+1} \sum_{i=r}^N  \text{var}(y_i)\label{varEP}.
\en
Our new ID estimation criterions make use of the updated variance on PDs:
$\text{var}(y_i)=\lambda_i-\hat{\sigma}^2. \label{e5}$

\begin{remark}\label{rm3.1}{\rm
Noise is typically different from outliers.
Noise affects every data points independently, while outliers are referred to
data points that are at least at a certain distance from the data points on manifold.
The proposed procedure is very robust to both noise and outliers, as shown in
experiments. On the other hand, the traditional PCA procedure can handle
limited noise but is very sensitive to outliers.
}
\end{remark}

\subsection{The local region selection method}

An embedding manifold can be approximated locally by linear subspaces.
The dimensionality of each linear subspace should be equal to the ID of
the embedding manifold. Therefore, it is possible to estimate the ID of a nonlinear
manifold by checking it locally. A cover is referred to a set whose elements
are subsets of the data set satisfying that the union of all subsets in the cover
contains the whole data set.

\begin{definition}[The set cover problem]
Given a universe $\mathcal{X}$ of $N$ elements and a collection $F$ of
subsets of $\mathcal{X}$, where $F=\{F_1,\cdots,F_N\}$. Set cover is concerned with
finding a minimum sub-collection of $F$ that covers all data points.
\end{definition}

Using a minimum cover has two advantages. First, it can find the minimal number of
subregions, which helps save the computational time. Secondly, the result of ID
estimation that utilizes the whole data set is more reliable. However,
searching such a minimal cover is an NP-hard problem. In the following,
we introduce an algorithm which can approximately find a minimal cover of a data set.

Given the parameter, an integer $k$ or a real number $\varepsilon$, there are
two ways to define the neighborhood of any data point $x$:
\begin{itemize}
\item[1)] The $k$-NN method: any data point $x_i$ that is one of $k$ nearest data points
of $x$ is in the neighborhood of $x$;
\item[2)] The $\varepsilon$-NN method: any data point $x_i$ in the region $\{y:
\|y-x\|<\varepsilon\}$ is in the neighborhood of $x$.
\end{itemize}

Without loss of generality, we may assume that the index of data points is independent of
their locations.

\begin{algorithm}\label{al3.1}\caption{\bf(Minimum set cover algorithm)}
{\bf Input:} Neighborhood size $k$ (integer) or $\varepsilon$ (real number),
             distance matrix $\hat{D}=(\|x_i-x_j\|)$\\
{\bf Output:} Minimum cover $F=\{(F_i,r_i),i=1,\cdots,S\}$.
\begin{algorithmic}[1]
\FOR{i=1 to $N$}
\STATE Identify the neighbors $\{x_{i_1},\cdots,x_{i_{P_i}}\}$ of $x_i$ by the $k$-NN or
$\varepsilon$-NN method. Let $F_i=\{i, i_1,\cdots,i_{P_i}\}$ be the index set of the
neighborhood and let $D$ be the $0-1$ incidence matrix.
\ENDFOR
\STATE Let $F=\{(F_i,r_i=0),i=1,\cdots,N\}$
\FOR{$i=1$ to $N$}
\STATE Let the frequency of $x_i$ be computed by
$Q_i=\sum_{j=1}^N D_{ij}$.
\ENDFOR
\FOR{$i=1$ to $N$}
 \IF{ $Q_i,Q_{i_1},\cdots,Q_{i_{P_i}} >1$}
 \STATE Remove $(F_i,r_i)$ from the cover set $F$ and set $Q_i=Q_i-1$,
$Q_{i_1}=Q_{i_1}-1$, $\cdots$, $Q_{i_{P_i}}=Q_{i_{P_i}}-1$.
 \ELSE
 \STATE Let $r_i=\underset{j=1,\cdots,P_i}\max\|x_i-x_{i_j}\|$
 \ENDIF
\ENDFOR
\end{algorithmic}

\end{algorithm}

Using the above approximation algorithm, a cover $F=\{(F_i,r_i),i=1,\cdots,S\}$ of
the data set $\mathcal{X}$ can be found. Compared with the local region selection
algorithm used in\cite{LocalPCA1971}, our algorithm above has a low time complexity
and avoids the supervised process to choose the neighborhood.
Intuitively, the cardinality $S$ of the cover $F$ satisfies that $N/k<S<N/2k$, where $k$
is the average number of neighbors.

\subsection{The proposed ID estimation algorithm}

We now present the proposed ID estimation algorithm using local PCA on the minimal
set cover: the C-PCD algorithms, which are summarized below for both batch and incremental
data, respectively.

\begin{algorithm}\label{al3.2}\caption{\bf(The C-PCA algorithm for batch data)}
\begin{description}
\item
{\bf Step 1.} Given a parameter $k$ or $\varepsilon$, compute a
minimal cover of $\mathcal{X}$ by Algorithm \ref{al3.1}.
Without loss of generality, $F=\{(F_i,r_i):\;i=1,\cdots,S\}$ is assumed to be the
constructed minimal set cover.

\item
{\bf Step 2.} Perform the PCA algorithm proposed in Subsections \ref{PCAdim}
and \ref{PCAdim2} on subsets $F_i$, $i=1\cdots,S$. The local ID estimations
$\{\hat{d}_i\}_{i=1}^S$ are then obtained.

\item
{\bf Step 3.} Let $\lambda_{ij}$ be the $j$-th eigenvalue on the $i$-th
subset in the decreasing order. $\lambda_j=\sum_i\lambda_{ij}$ is
considered as the variance of $\mathcal{X}$ on its $j$-th PD.
Subsequently, the global ID estimation $\hat{d}$ can be derived using the
criterions (\ref{e2}) or (\ref{e3}).
\end{description}
\end{algorithm}

In many cases, consecutive data are collected incrementally. This
requires an incremental learning algorithm to inspect the change of the
data structure on time. The incremental C-PCA algorithm is presented
as follows.

\begin{algorithm}\label{al3.3}\caption{\bf(The incremental C-PCA algorithm)}
\begin{description}
\item
{\bf Step 1.} The new data point is assumed to be $x$. Let $\{x_1,\cdots,x_S\}$
be the centers of the subsets in the cover. Find the nearest center $x_q$ of $x$:
$x_q=\arg\underset{i=1,\cdots,S}{\min}\|x-x_i\|$.

\item
{\bf Step 2.} If $\|x_q-x\|>r_q$, then the data point $x$ is considered as an
outlier and the remaining part of the algorithm will not be performed on $x$.
Otherwise, go to {\bf Step 3}.

\item
{\bf Step 3.} Performs PCA on $F_q=F_q\bigcup\{x\}$. Let $\lambda_{qj}'$ be the $j$-th
eigenvalue. Update $\lambda_j$ by $\lambda_j=\lambda_j+\lambda_{qj}'-\lambda_{qj}$.
Then let $\lambda_{qj}=\lambda_{qj}'$.

\item
{\bf Step 4.} Update the local ID, $\hat{d}_q$, and the global ID, $\hat{d}$,
of $\mathcal{X}$.
\end{description}
\end{algorithm}

\begin{remark}\label{rm3.2}{\rm
Our method is different from the Local-PCA \cite{LocalPCA1971} in many aspects.
First, the centers and the local regions are determined simultaneously by using one
parameter - the neighborhood size, whilst, in \cite{LocalPCA1971}, the centers
and neighborhood sizes are determined by two parameters. Secondly, our approach finds
the subregions by approximating a minimum cover of the data set,
while the local-PCA in \cite{LocalPCA1971} does not guarantee whether or not
the selected subregions cover the whole data set.
}
\end{remark}

\subsection{Computational complexity analysis}

The computational complexity of our algorithms is one of the most
important issues for its application. The batch mode ID estimation
can be divided into two parts. In the first part, computing the distance
matrix needs $O(N^2)$ time, searching the nearest neighbors for every
data point needs $O(kN^2)$ time and finding an approximate minimum cover of
$\mathcal{X}$ needs $O(kN)$ time. Therefore, the first part needs
$O((K+1)N^2+kN)$ running time. In the second part, performing PCA
locally needs $k^3\times({N}/{k})\approx O(k^2N)$ running
time. To sum up, the total running time needed for the batch mode
algorithm is $O((k+1)N^2 +(k^2+k)N)$. If the proposed method is
embedded in a manifold learning algorithm, then the running time complexity can
be reduced to $O((k^2+k)N)$ in the case when the distance matrix and the
neighborhood are already defined. This is a relatively small increase in the
time complexity of a manifold learning algorithm which is always as high as $O(N^3)$.

For incremental learning, the neighborhood identification step needs
$O(N/k)$ running time, whilst the local PCA consumes $O((k+1)^3)$ running time.
Therefore, the total time complexity for incremental learning is $O((k+1)^3+N/k)$.

\section{Experiments}\label{exp}

The proposed algorithm was implemented with parameters $\alpha=10$ and $\beta=0.8$
for all the experiments.

In practice, it is found that noise contained in data is of low-dimension, except
an additive white noise which is assumed to be in every component of the data
vectors in $\R^D$.
Thus, in practice, we only use variances of the first $\min(10,N-r+1)$ PCs
in the noise part of data to estimate the variance of noise (see Eq. (\ref{varEP})).

Comparison is made among the $k$-$k/2$ NN method \cite{manifoldAdapt},
the $k$-NNG method \cite{GeodesicEntropic2004}, the revised MLE (MLE in short)
method \cite{revisedMLE}, the C-PCA method and the L-PCA method,
where the L-PCA method stands for the C-PCA method without
the noise filtering procedure proposed in Subsection \ref{PCAdim2}.
It should be noted that the results obtained by the MLE,
$k$-$k/2$ NN and $k$-NNG methods are positive real numbers,
while the L-PCA and C-PCA methods produce only integer ID estimations.
In order to make a comparison among these results,
we average the local ID estimations obtained
with the C-PCA and L-PCA methods to provide a real ID estimation:
$\hat{d}=\frac{1}{S}\sum_{i=1}^S\hat{d}_i.$

\subsection{$10$-Mobius data}

\begin{figure}
\includegraphics[width=6cm]{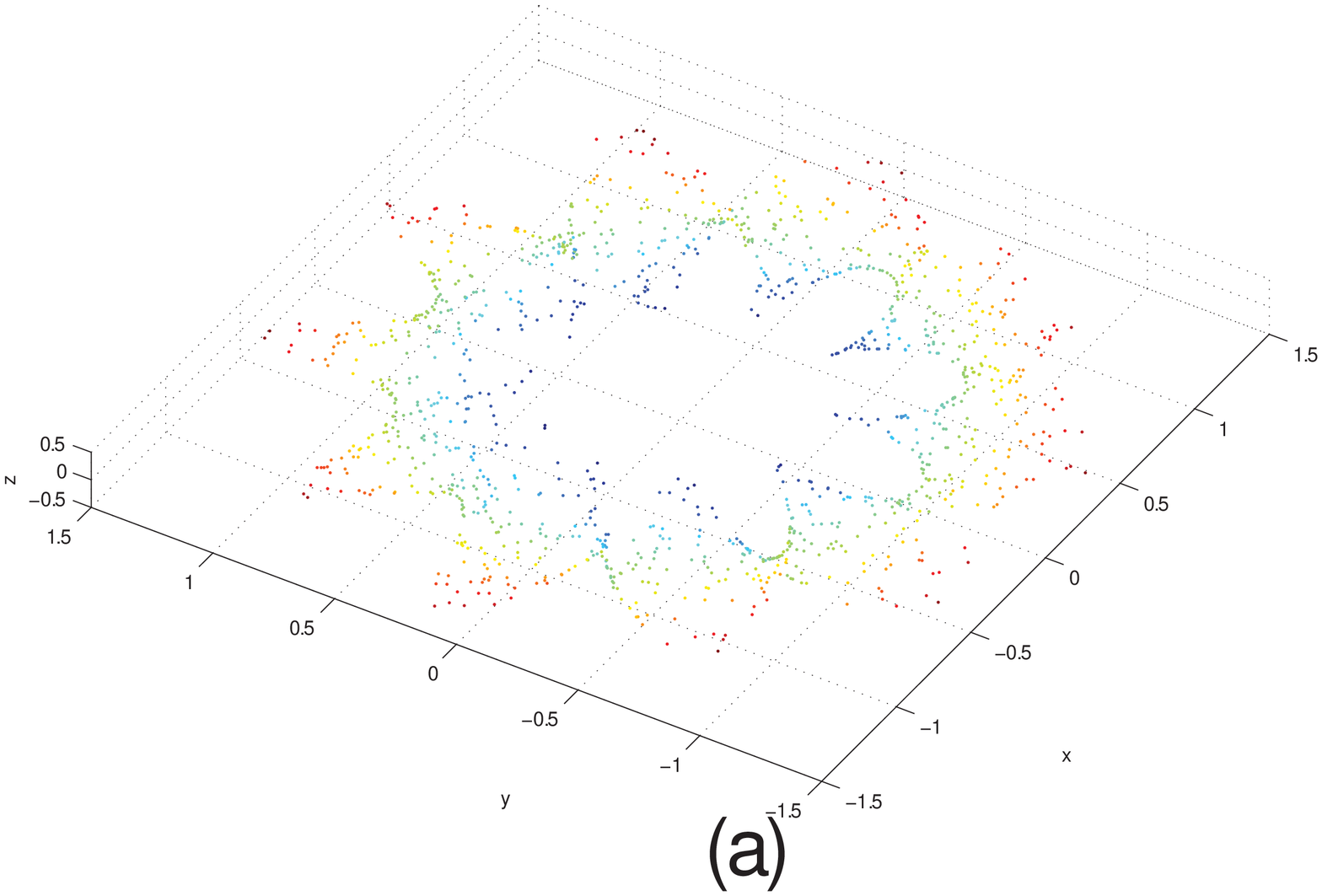}
\includegraphics[width=8cm]{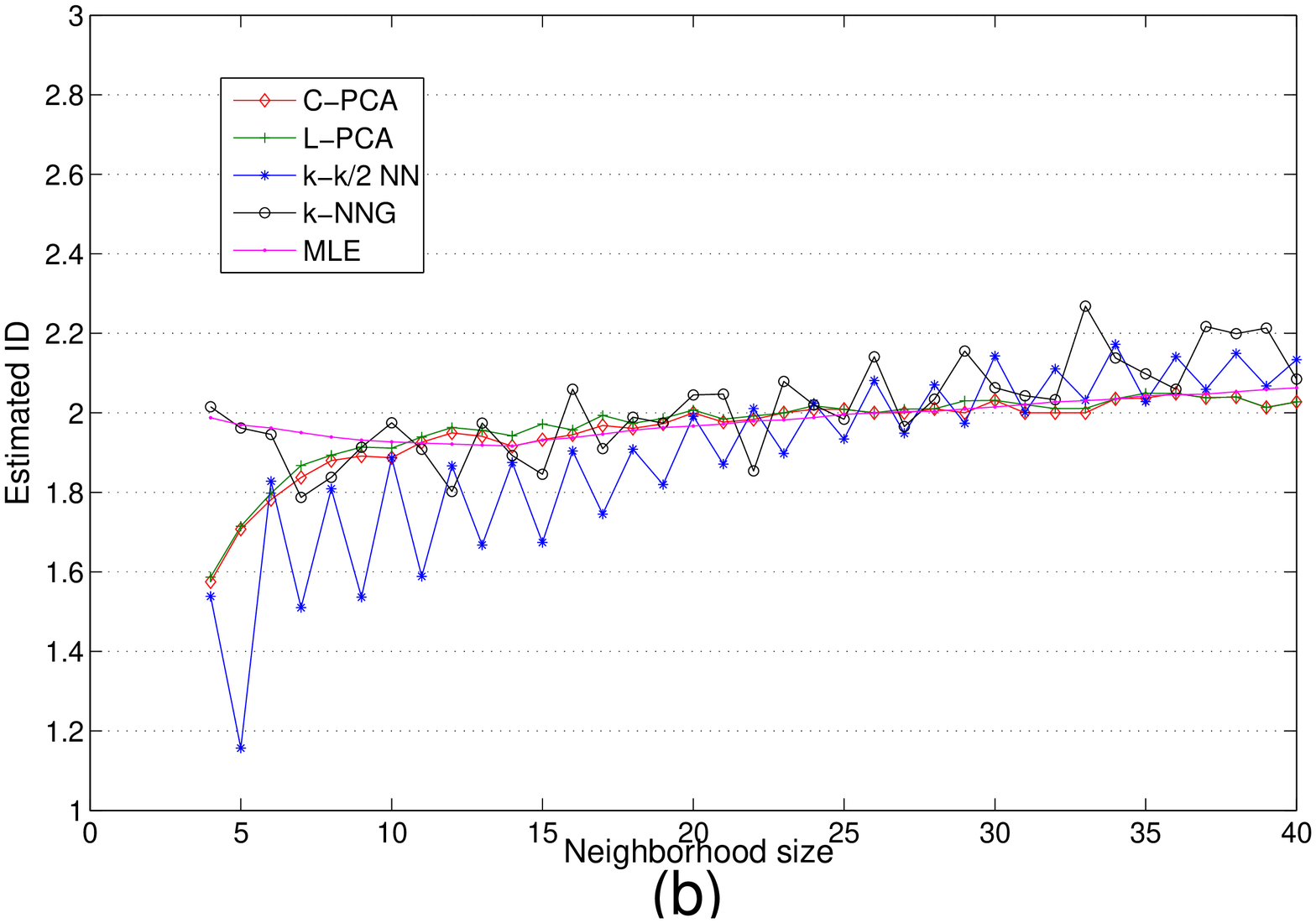}
\caption{(a) shows the scatter plot of the Mobius ring data set, and (b)
shows the ID estimation results corresponding to the size of subregions.} \label{mobius}
\end{figure}

The first data set is a $10$-Mobius ring embedded in $\R^3$. Fig. \ref{mobius}(a) shows the
scatter plot of the Mobius ring data set. As can be seen, the Mobius data points are lying
on a highly nonlinear manifold with $1200$ points uniformly distributing on the surface.
Fig. \ref{mobius}(b) shows the results obtained by the five ID estimation algorithms
against the neighborhood size ranging from $4$ to $40$. The MLE method is the most stable
and accurate algorithm for all neighborhood sizes. All algorithms converge to the correct
estimation. It seems that the L-PCA method does not diverge on this data set. This is possibly
because the original dimensionality of data is low.

\subsection{Real world data sets}

\begin{figure*}
\includegraphics[width=6cm]{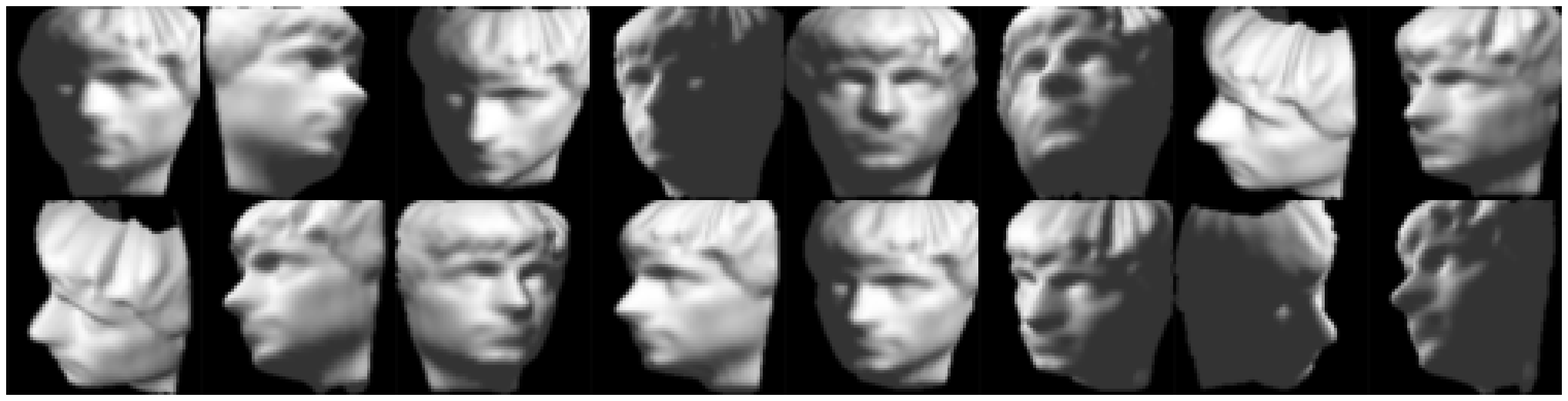}
\includegraphics[width=8cm]{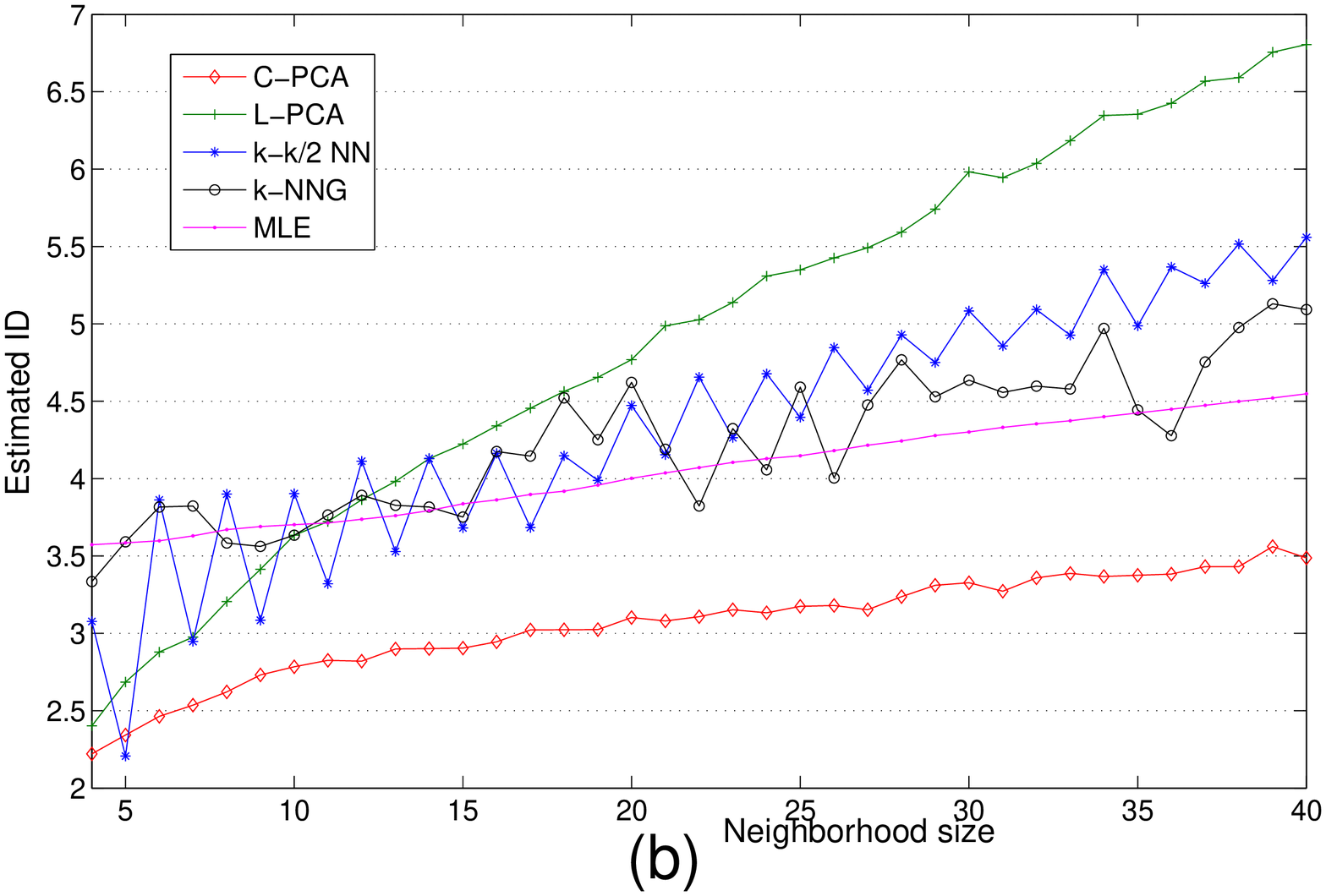}
\caption{(a) shows some samples of the Isoface data set. As can be seen, a head is under
left-right, up-down and lighting changes. (b) presents the estimated ID of the Isoface data set.
}\label{isofaces}
\end{figure*}

\begin{figure*}
\includegraphics[width=6cm]{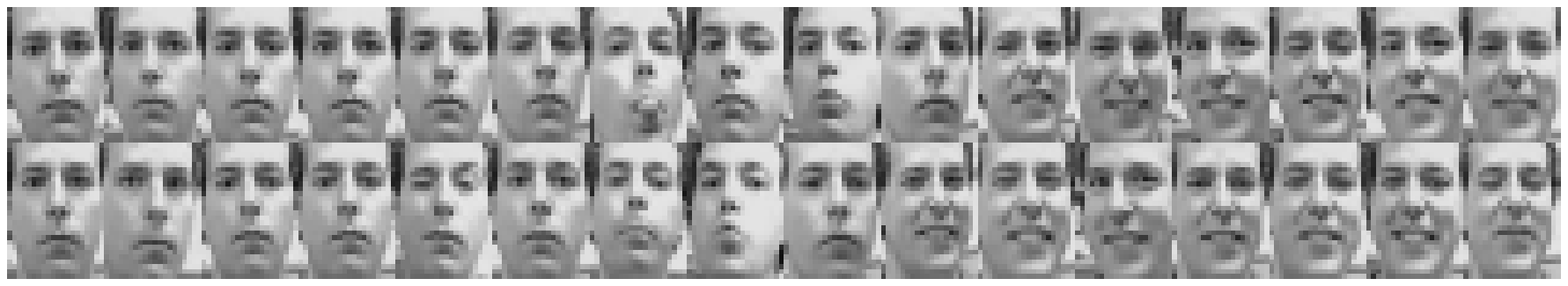}
\includegraphics[width=8cm]{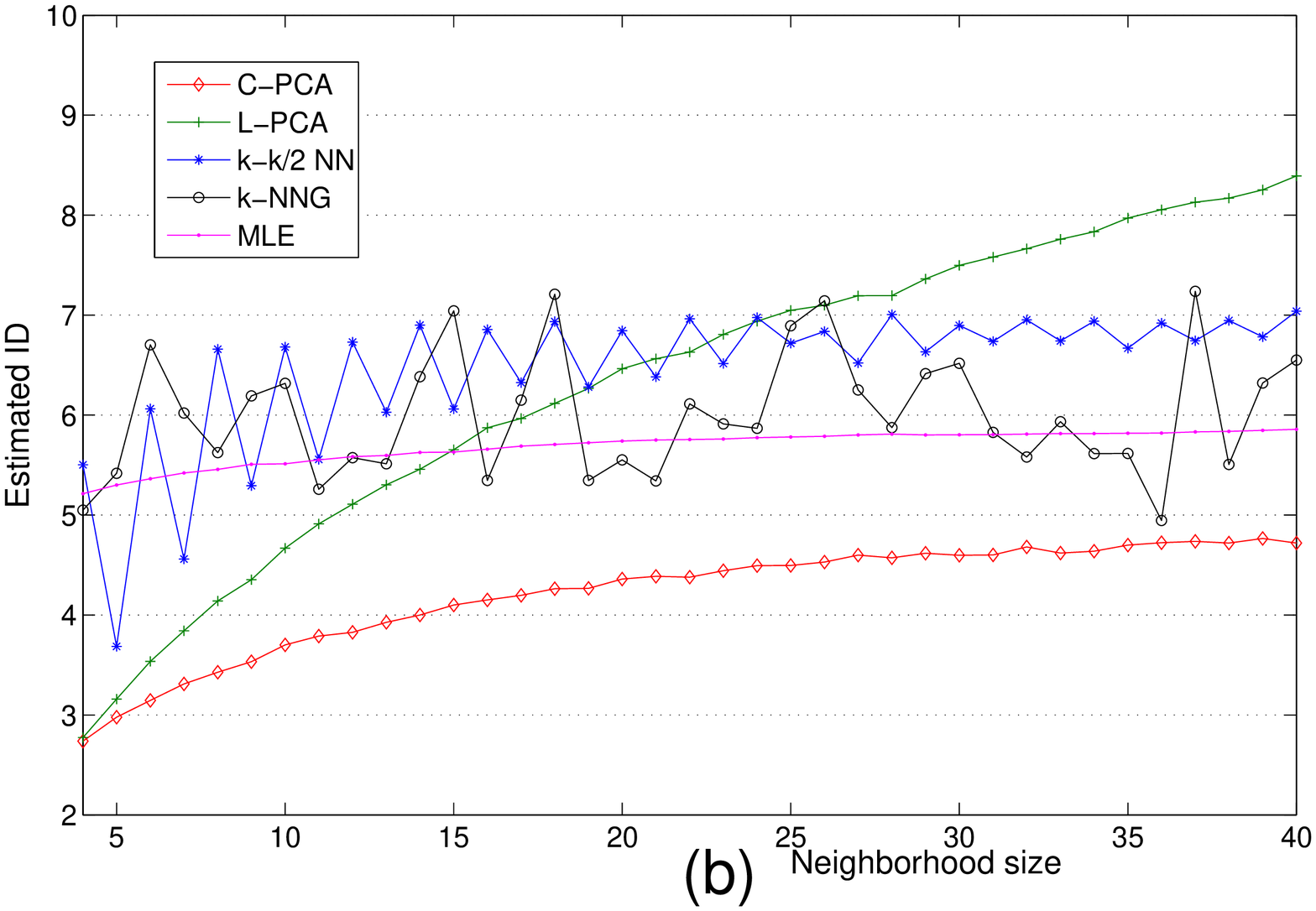}
\caption{(a) shows some samples of the LLEface data set and (b) plots the estimated
ID of the LLEface data set against the neighborhood size.}
\label{LLEfaces}
\end{figure*}

\begin{figure*}
\includegraphics[width=6cm]{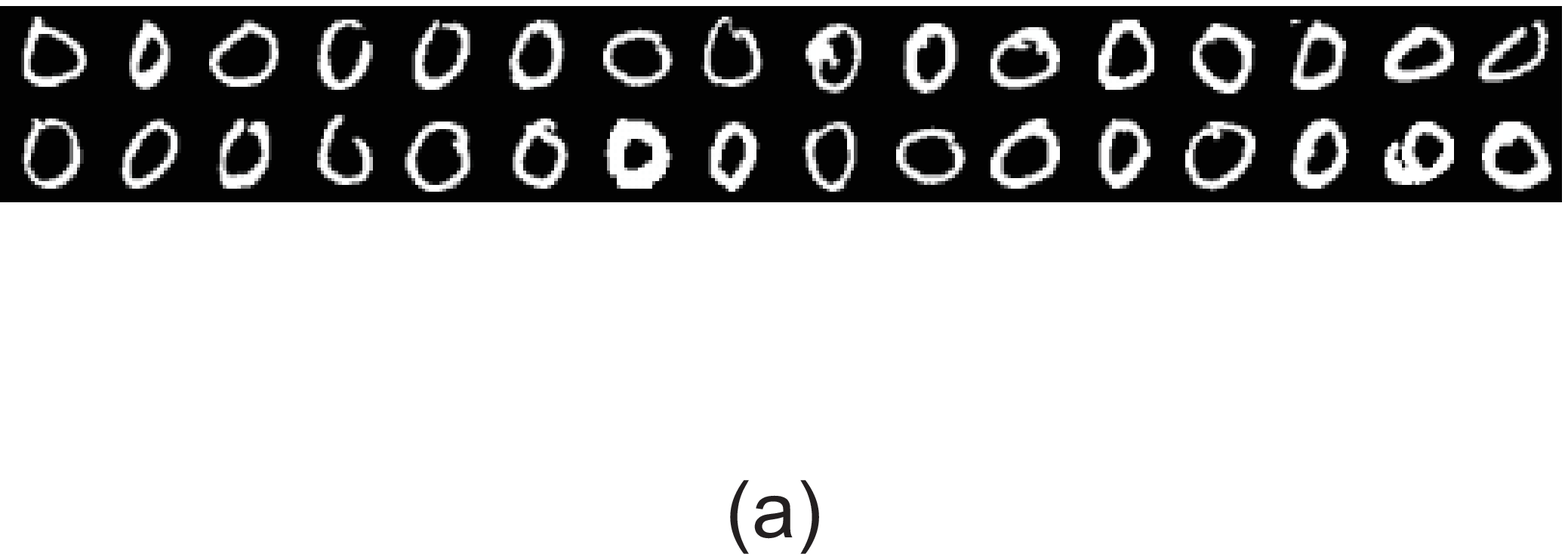}
\includegraphics[width=8cm]{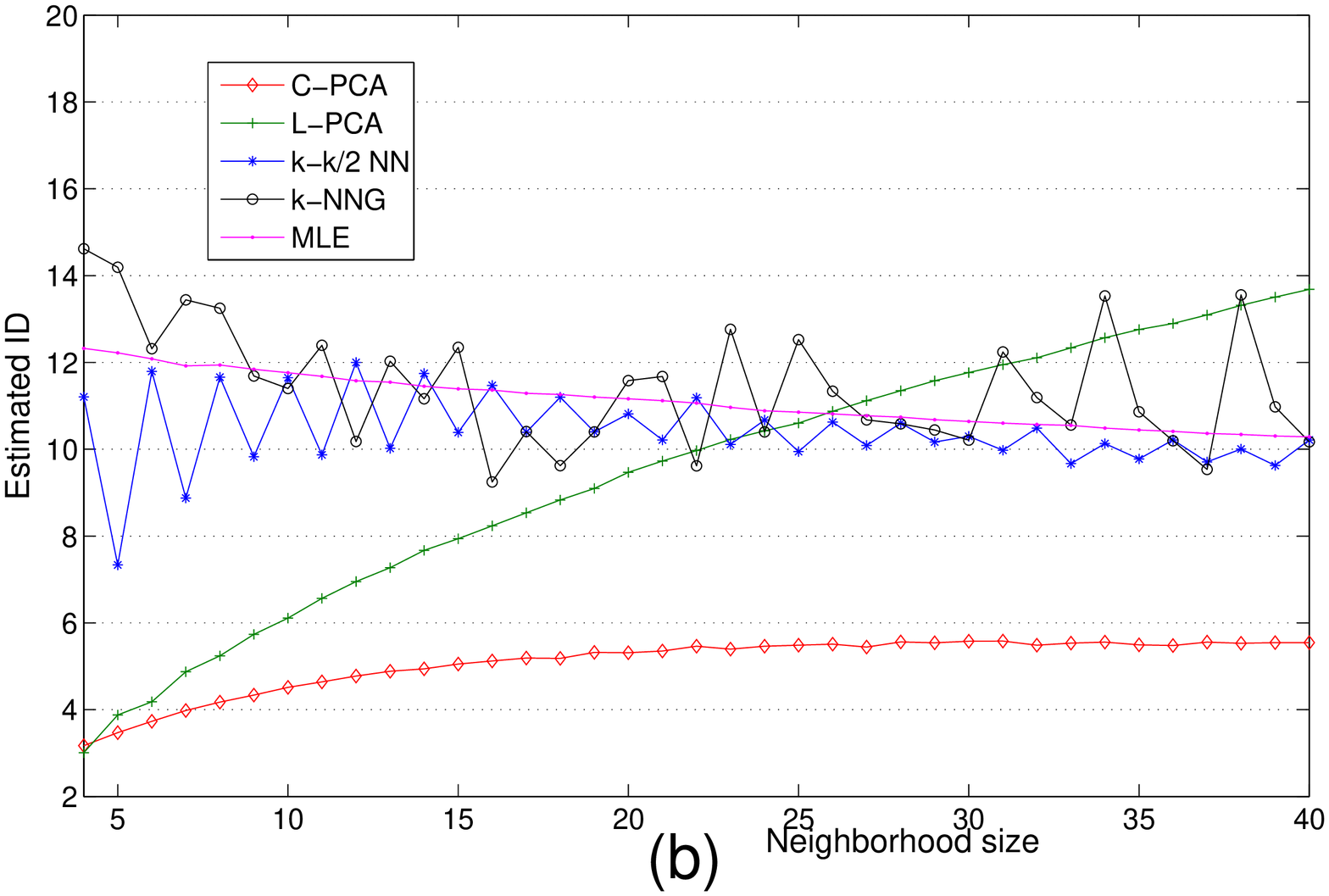}
\caption{(a) shows some samples of '$0$' in the MNIST data set and (b) gives the plot of
the estimated ID of data '$0$' versus the neighborhood size.}
\label{MNIST0}
\end{figure*}

\begin{figure*}
\includegraphics[width=6cm]{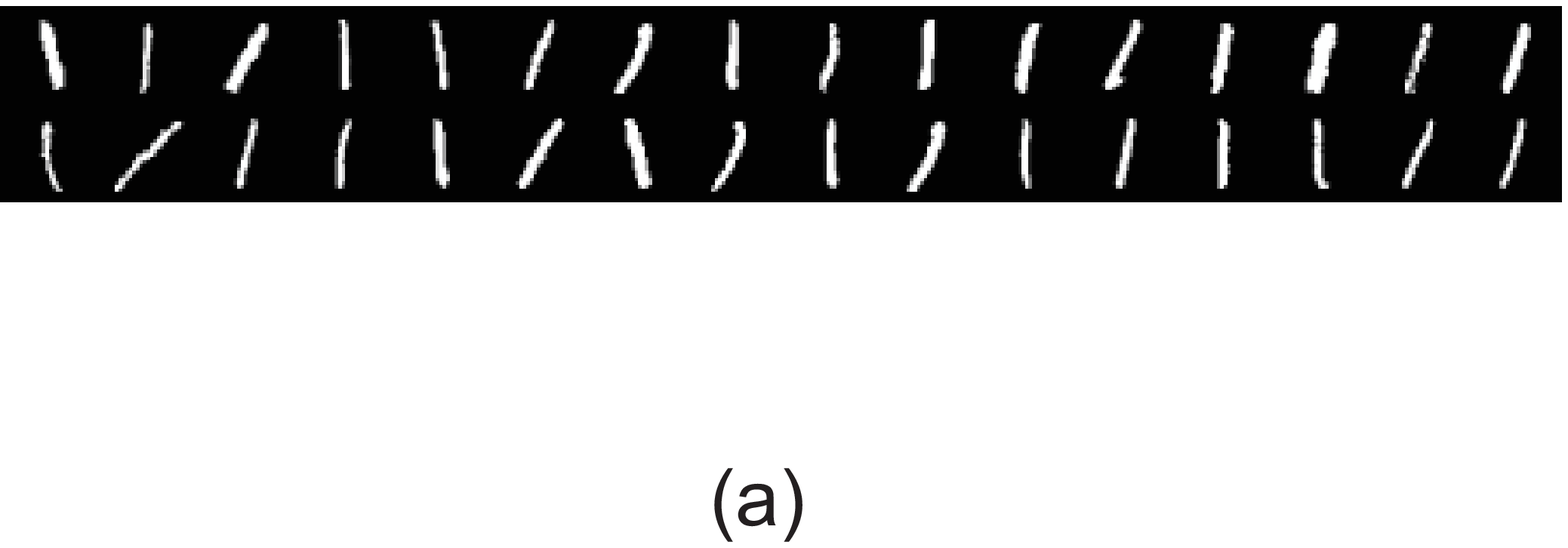}
\includegraphics[width=8cm]{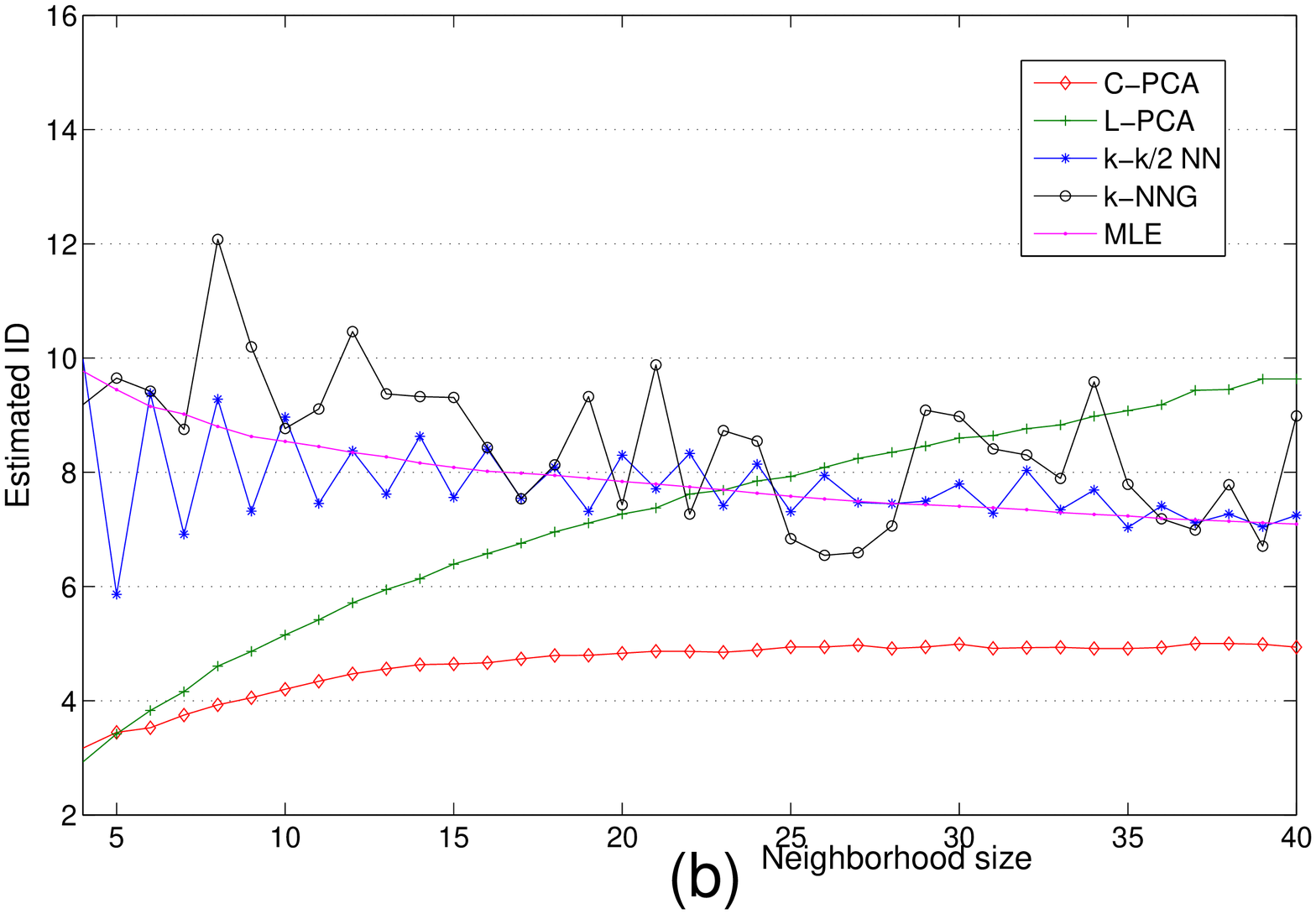}
\caption{(a) shows some samples of '$1$' in the MNIST data set and (b) presents the
plot of the estimated ID of data '$1$' versus the neighborhood size.}
\label{MNIST1}
\end{figure*}

Our algorithm is compared with the MLE, $k$-$k/2$ NN and $k$-NNG methods on some
benchmark real world data sets: the Isoface data set \cite{T2000},
the LLEface data set \cite{R2000} and the MNIST '$0$' and '$1$' data sets \cite{MNIST}.

The Isoface data set is comprised of $698$ images of a head with the resolution
$64\times 64$. Some samples of the Isoface data set are shown in Fig. \ref{isofaces}(a).
In the experiments, each image is reshaped to a $4096$-dimensional vector.
It can be seen that the Isoface data set is under a three-dimensional movement:
up-down, left-right and lighting changes. In \cite{T2000}, the Isomap
algorithm estimated its ID as $3$ using the projection approach.
As can be seen from Fig. \ref{isofaces}(b), corresponding to the neighborhood sizes
from $4$ to $40$, the C-PCA estimator ranges from $2.3$ to $3.5$ and the MLE estimator
ranges from $3.5$ to $4.5$. The estimation given by the $k$-NNG and $k$-$k/2$ NN methods
is oscillating badly with the neighborhood sizes, so they are bot unstable.
Since the L-PCA method can not filter out noise contained in data, it tends to overestimate
the ID as the neighborhood size increases. This means that our noise filtering process
plays a key role in the convergence of the C-PCA method.

The second data set is the LLEface data set, which contains $1965$ samples in a
$560$-dimensional space (see Fig. \ref{LLEfaces}(a) for some samples).
From Fig. \ref{LLEfaces}(b), it is seen that both the C-PCA and the MLE methods give
a convergent ID estimation with the increase of the neighborhood size,
while the L-PCA, $k$-$k/2$-NN and $k$-NNG methods seem not convergent when
the neighborhood size is increasing.
The ID estimation given by the C-PCA method changes between $2.8$ and $4.7$
with the convergent estimation being $4.7$, while the estimation result
obtained by the MLE method changes gradually from $5.2$ to $5.8$ with
a convergent estimation of $5.8$.

We now consider two MNIST data sets: the set '0' and the set '1' (see Fig. \ref{MNIST0}(a)
and Fig. \ref{MNIST1}(a) for some samples of these two data, respectively). The data set '0'
contains $980$ data points, while the data set '1' contains $1135$ data points.
It can be seen from Fig. \ref{MNIST0}(b) and Fig. \ref{MNIST1}(b) that
all methods, except the L-PCA and $k$-NNG methods, converge with the increase of the
neighborhood size.
For the data set '0', it can be seen from Fig. \ref{MNIST0}(b) that the ID estimation
given by the C-PCA method converges to $5.8$ and the estimation given by both the MLE
estimator and the $k$-$k/2$-NN estimator converges to $10$. For the data set '1',
Fig. \ref{MNIST1}(b) shows that the ID estimation obtained by the C-PCA method converges
to $5.5$ and the estimation provided with both the MLE method and the $k$-$k/2$-NN
method converges to $7.2$.
Note that the result given by our method is in a big disagreement with the results given by
other methods for the ID estimation of the data sets '$0$' and '$1$'.
A digit '$0$' is usually represented as an ellipse which can be determined by the coordinates
of its focus and its major and minor axes, so the ID of the data set '$0$' is likely to be $5$.
The number '$1$' can be considered as a line segment, which rotates from left to
right, so a sensible ID estimation for the data set '$1$' may be between $4$ and $5$.

\subsection{Noisy data sets}

\begin{figure}
\centering
\includegraphics[width=6cm]{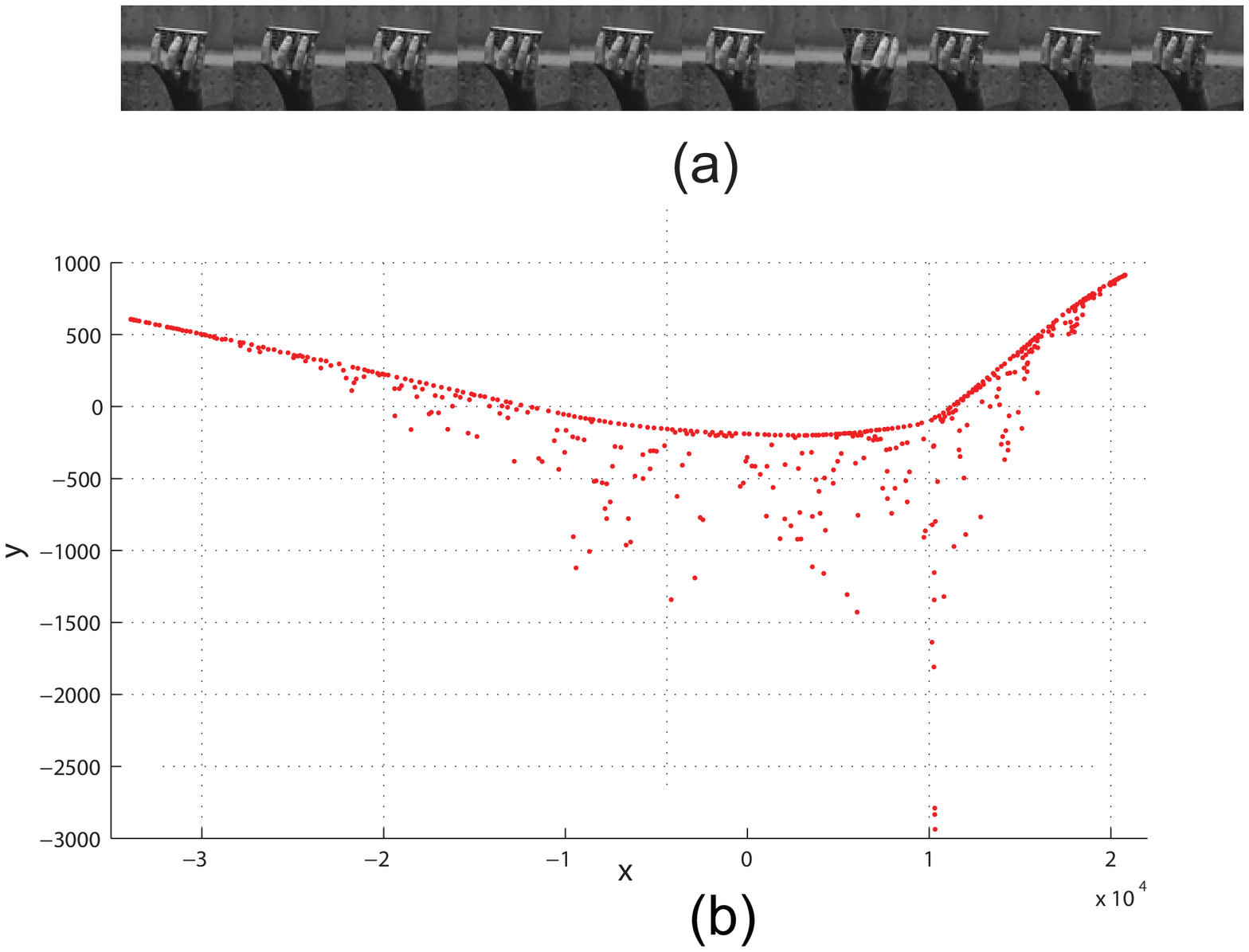}
\includegraphics[width=7cm]{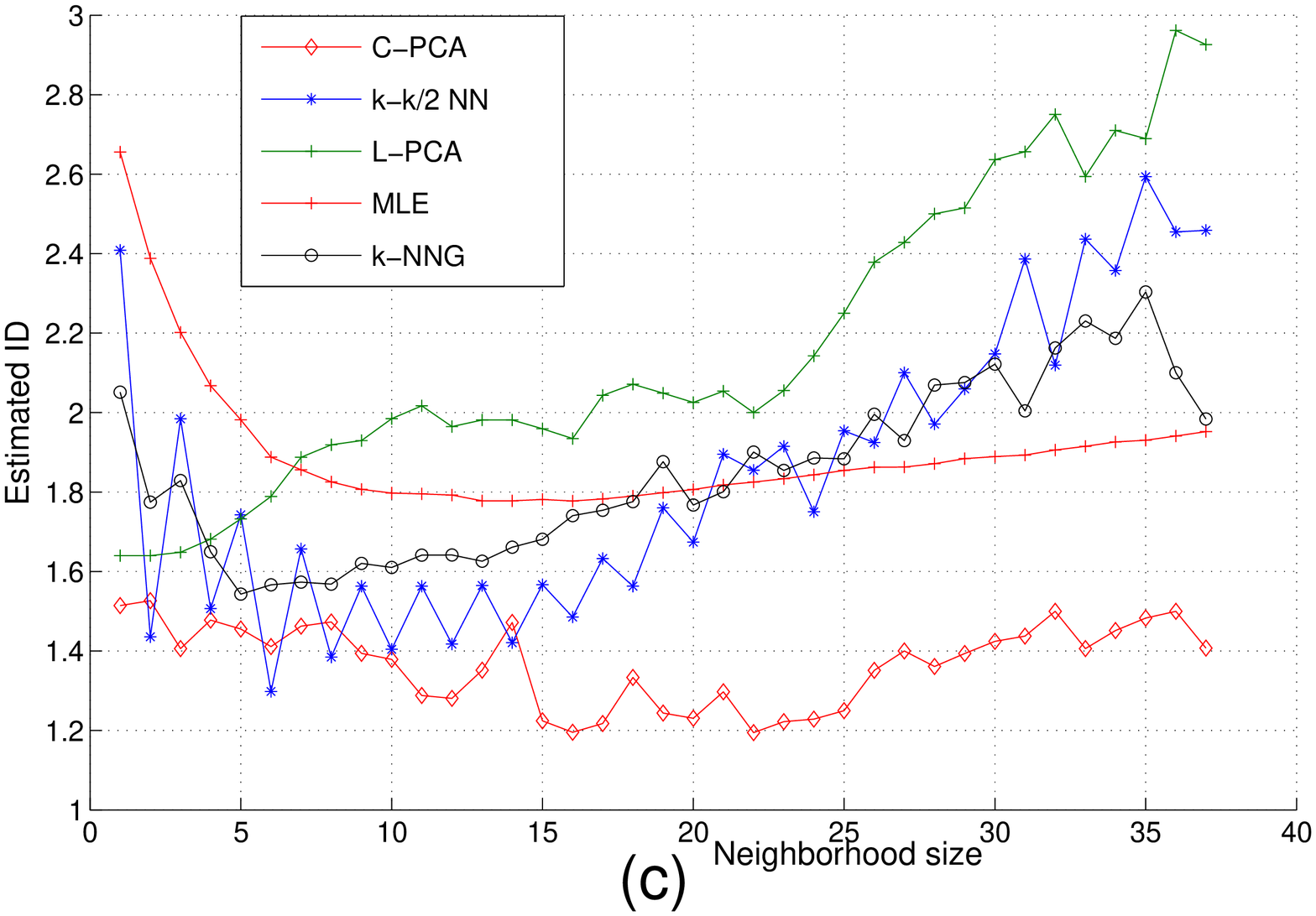}
\caption{(a) shows selected samples of the hand rotation data set,
(b) shows the low dimensional embedding of hands rotation data sets
by Isomap algorithm, (c) ID estimations of the hands rotation data set.
}\label{hands}
\end{figure}

\begin{figure}
\centering
\includegraphics[width=10cm]{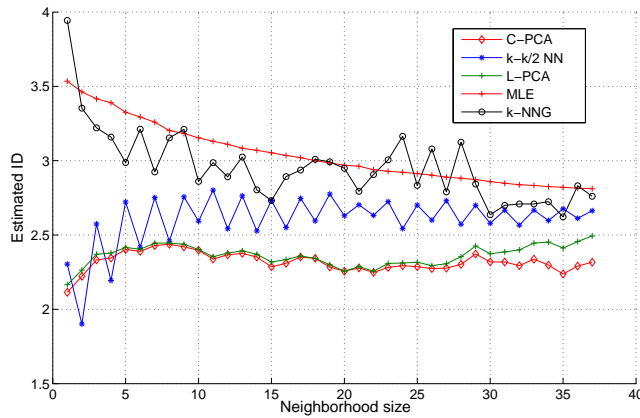}
\caption{ID estimations of the noised Mobius data set.
}\label{MobiusNoise2}
\end{figure}

The traditional PCA algorithm is very sensitive to outliers, and the performance of PCA-based
algorithms deteriorate rapidly if data points are sparse on a manifold such as the hand rotation
data set \footnote{CMU database: http://vasc.vi.cmu.edu/idb/html/motion/hand/index.html}.
As can be seen from Fig. \ref{hands}(a), the hand is under a one-dimensional movement,
so the data points can be considered as lying on a one-dimensional curve.
The data set contains $481$ image samples, and each sample is a vector in a
$512480$-dimensional space. Many outliers can be seen from its low-dimensional embedding by
the Isomap algorithm (see Fig. \ref{hands}(b)). Its ID estimation results with different
methods are shown in Fig. \ref{hands}(c).

Both the $k$-$k/2$ NN and $k$-NNG methods are sensitive to the choice of the
neighborhood size and tend to overestimate the ID as the neighborhood
size increases. On the other hand, the MLE estimator is more stable
(see Fig. \ref{hands}(c)). However, the minimum estimation of MLE
method is $1.75$, which is still higher than the ID of this data set.
L-PCA method has the worst performance due to the outliers contained in the data set.
The estimation of the C-PCA method, which changes between $1.5$ and $1.2$, is the closest
one to the correct ID of this data set.

We now transform the original $10$-Mobius data in a 4-dimensional space
using an Euclidean transformation. A random noise with mean $0$ and variance $0.2$ is
added to the transformed data. The ID estimation results with different algorithms
are given in Fig. \ref{MobiusNoise2}. As can be seen from Fig. \ref{MobiusNoise2},
the ID estimation given by the C-PCA method is the closest one to the correct ID of
this noised $10$-Mobius data set. The other algorithms tend to overestimate the ID of
the noised data set. The estimation obtained by the L-PCA method is a little higher
than that given by the C-PCA due to the effect of noise.

\section{Conclusion}\label{con}

In this paper, we proposed a new ID estimation method based on PCA.
The proposed algorithm is simple to implement and gives a convergent
ID estimation corresponding to a wide range of neighborhood sizes.
It is also convenient for incremental learning. Experiments have shown that
the new algorithm has a robust performance.

\section*{Acknowledgment}

The work of H. Qiao was supported in part by the National Natural
Science Foundation (NNSF) of China under grant no. 60675039 and 60621001
and by the Outstanding Youth Fund of the NNSF of China under grant no. 60725310.
The work of B. Zhang was supported in part by the 863 Program of China under
grant no. 2007AA04Z228, by the 973 Program of China under grant no. 2007CB311002
and by the NNSF of China under grant no. 90820007.

\end{document}